\documentclass[journal]{IEEEtran}
\usepackage{cite}
\usepackage{multirow}
\usepackage{array}
\usepackage{amsmath,graphicx,float}
\usepackage[hidelinks]{hyperref}
\usepackage{commath}
\usepackage{afterpage}
\usepackage{xcolor}
\usepackage[ruled,vlined]{algorithm2e}
\usepackage{booktabs}
\usepackage{caption}
\usepackage[utf8]{inputenc}
\graphicspath{{figure/}}

\begin{document}
\title{A Tandem Learning Rule for Effective Training and Rapid Inference of Deep Spiking Neural Networks}

\author{Jibin~Wu,
	Yansong~Chua,
	Malu~Zhang,
	Guoqi~Li,
	Haizhou~Li,~
	and~Kay~Chen~Tan~
\thanks{J.~Wu, M. Zhang and H. Li are with the Department
	of Electrical and Computer Engineering, National University of Singapore, 
	(e-mail: jibin.wu@u.nus.edu, maluzhang@nus.edu.sg, haizhou.li@nus.edu.sg).}
\thanks{Y.~Chua is with the Institute for Infocomm Research, A*STAR, Singapore,
	(Corresponding author, e-mail: james4424@gmail.com.)}
\thanks{G.~Li is with the Center for Brain Inspired Computing Research and Beijing Innovation Center for Future Chip, Department of Precision Instrument, Tsinghua University, P. R. China., (e-mail: liguoqi@mail.tsinghua.edu.cn)}
\thanks{K.~C.~Tan is with the Department of Computer Science,
		City University of Hong Kong, Hong Kong,
		(e-mail: kaytan@cityu.edu.hk).}
}

\maketitle
\SetKwInput{KwInput}{Input}                
\SetKwInput{KwOutput}{Output}              
\SetKwInput{KwFw}{Forward Pass}       
\SetKwInput{KwBw}{Backward Pass}       
\SetKwInput{KwLoss}{Loss}       
\SetKwInput{KwNote}{Note}       

\begin{abstract}

Spiking neural networks (SNNs) represent the most prominent biologically inspired computing model for neuromorphic computing (NC) architectures. However, due to the non-differentiable nature of spiking neuronal functions, the standard error back-propagation algorithm is not directly applicable to SNNs. In this work, we propose a tandem learning framework, that consists of an SNN and an Artificial Neural Network (ANN) coupled through weight sharing. The ANN is an auxiliary structure that facilitates the error back-propagation for the training of the SNN at the spike-train level. To this end, we consider the spike count as the discrete neural representation in the SNN, and design ANN neuronal activation function that can effectively approximate the spike count of the coupled SNN. The proposed tandem learning rule demonstrates competitive pattern recognition and regression capabilities on both the conventional frame-based and event-based vision datasets, with at least an order of magnitude reduced inference time and total synaptic operations over other state-of-the-art SNN implementations. Therefore, the proposed tandem learning rule offers a novel solution to training efficient, low latency, and high accuracy deep SNNs with low computing resources.
\end{abstract}


\section{INTRODUCTION}
\label{intro}
Deep learning has greatly improved pattern recognition performance by leaps and bounds in computer vision \cite{krizhevsky2012imagenet}, speech processing \cite{xiong2017toward}, language understanding \cite{hirschberg2015advances} and robotics \cite{silver2017mastering}. However, deep artificial neural networks (ANNs) are computationally intensive and memory inefficient, thereby, limiting their deployments in mobile and wearable devices that have limited computational budgets. This prompts us to look into energy-efficient solutions.

The human brain, with millions of years of evolution, is incredibly efficient at performing complex perceptual and cognitive tasks. Although hierarchically organized deep ANNs are brain-inspired, they differ significantly from the biological brain in many ways. Fundamentally, the information is represented and communicated through asynchronous action potentials or spikes in the brain. To efficiently and rapidly process the information carried by these spike trains, biological neural systems evolve the event-driven computation strategy, whereby energy consumption matches with the activity level of sensory stimuli. 

Neuromorphic computing (NC), as an emerging non-von Neumann computing paradigm, aims to mimic such asynchronous event-driven information processing with spiking neural networks (SNNs) in silicon \cite{schuman2017survey}. The novel neuromorphic computing architectures, instances include TrueNorth \cite{merolla2014million} and Loihi \cite{davies2018loihi}, leverage on the low-power, densely-connected parallel computing units to support spike-based computation. Furthermore, the co-located memory and computation can effectively mitigate the problem of low bandwidth between the CPU and memory (i.e., von Neumann bottleneck) \cite{monroe2014neuromorphic}. When implemented on these neuromorphic architectures, deep SNNs benefit from the best of two worlds: superior classification accuracies and compelling energy efficiency \cite{Esser11441}. 

While neuromorphic computing architectures offer attractive energy-saving, how to train large-scale SNNs that can operate efficiently and effectively on these NC architectures remains a challenging research topic. The spiking neurons exhibit a rich repertoire of dynamical behaviours\cite{izhikevich2004model}, such as phasic spiking, bursting, and spike frequency adaptation, which significantly increase the modeling complexity over the simplified ANNs. Moreover, due to the asynchronous and discontinuous nature of synaptic operations within the SNN, the error back-propagation algorithm that is commonly used for the ANN training is not directly applicable to the SNN.

Over the years, a growing number of neural plasticities or learning methods, inspired by neuroscience and machine learning studies, have been proposed for SNNs \cite{pfeiffer2018deep, tavanaei2018deep}. The biological plausible Hebbian learning rules \cite{hebb2005organization} and spike-timing-dependent plasticity (STDP) \cite{bi1998synaptic} are intriguing local learning rules for computational neuroscience studies and also attractive for hardware implementation with emerging non-volatile memory device \cite{burr2017neuromorphic}. Despite their recent successes on the small-scale image recognition tasks \cite{mozafari2018first,kheradpisheh2018stdp}, they are not straightforward to be used for large-scale machine learning tasks due to the ineffective task-specific credit assignment and time-consuming hyperparameter tuning.  

Recent studies \cite{diehl2015fast,ethImageNet,sengupta2019going} show that it is viable to convert a pre-trained ANN to an SNN with little adverse impacts on classification accuracy. This indirect training approach assumes that the activation value of analog neurons is equivalent to the average firing rate of spiking neurons, and simply requires parsing and normalizing of weights of the trained ANN. Rueckauer et al. \cite{ethImageNet} provide a theoretical analysis of the performance deviation of such an approach as well as a systematic study on the Convolutional Neural Network (CNN) models for the object recognition task. This conversion approach achieves the best-reported results for SNNs on many conventional frame-based vision datasets including the challenging ImageNet-12 dataset \cite{ethImageNet, sengupta2019going}. However, this generic conversion approach comes with a trade-off that has an impact on the inference speed and classification accuracy and requires at least several hundred of inference time steps to reach an optimal classification accuracy. 

Additional research efforts are also devoted to training constrained ANNs that can approximate the properties of SNNs \cite{hunsberger2016training, wu2019deep}, which allow the trained model to be transferred to the target hardware platform seamlessly. Grounded on the rate-based spiking neuron model, this constrain-then-train approach transforms the steady-state firing rate of spiking neurons into a continuous and hence differentiable form that can be optimized with the conventional error back-propagation algorithm. By explicitly approximating the properties of SNNs during the training process, this approach performs better than the aforementioned generic conversion approach when implemented on the target neuromorphic hardware. 

While competitive classification accuracies are shown with both the generic ANN-to-SNN conversion and the constrain-then-train approaches, the underlying assumption of a rate-based spiking neuron model requires a long encoding time window (i.e., how many time steps the image or sample are presented) or a high firing rate to reach the steady neuronal firing state\cite{ethImageNet, hunsberger2016training}, such that the approximation errors between the pre-trained ANN and the SNN can be eliminated. This steady-state requirement limits the computational benefits that can be acquired from the NC architectures and remain a major roadblock for applying these methods to real-time pattern recognition tasks. 

To improve the overall energy efficiency as well as inference speed, an ideal SNN learning rule should support a short encoding time window with sparse synaptic activities. To exploit this desirable property, the temporal coding has been investigated whereby the spike timing of the first spike was employed as a differentiable proxy to enable the error back-propagation algorithm \cite{mostafa2018supervised, hong2019training, zhang2020spike}. Despite competitive results on the MNIST dataset, it remains elusive how the temporal learning rule maintains the stability of neuronal firing such that the derivatives can be determined, and how it can be scaled up to the size of state-of-the-art deep ANNs. In view of the steady-state requirement of rate-based SNNs and scalability issues of temporal-based SNNs, it is necessary to develop new learning methods that can effectively and efficiently train deep SNNs to operate under short encoding time window with sparse synaptic activities.  

Surrogate gradient learning \cite{neftci2019surrogate} has emerged recently as an alternative training method for deep SNNs. With a discrete-time formulation, the spiking neuron can be effectively modeled as a non-spiking recurrent neural network (RNN), wherein the leak term in spiking neuron models is formulated as a fixed-weight self-recurrent connection. By establishing the equivalence with RNNs, the error Back-propagation Through Time (BPTT) algorithm can be applied to train deep SNNs. The non-differentiable spike generation function can be replaced with a continuous function during the error back-propagation, whereby a surrogate gradient can be derived based on the instantaneous membrane potential at each time step. In practice, the surrogate gradient learning performs exceedingly well for both static and temporal pattern recognition tasks \cite{lee2016training, shrestha2018slayer, wu2018direct, zenke2018superspike}. By removing the constraints of steady-state firing rate for rate-based SNN and spike-timing dependency of temporal-based SNN, the surrogate gradient learning supports rapid and efficient pattern recognition with SNNs.

While competitive accuracies were reported on the MNIST and CIFAR-10 \cite{krizhevsky2009learning} datasets with the surrogate gradient learning, it is both memory and computationally inefficient to train deep SNNs using BPTT, especially for more complex datasets and network structures. Furthermore, the vanishing gradient problem \cite{hochreiter1998vanishing} that is well-known for vanilla RNNs may adversely affect the learning performance for spiking patterns with long temporal duration. In this paper, to improve the learning efficiency of the surrogate gradient learning, we propose a novel learning rule with the tandem neural network. As illustrated in Fig.\ref{fig:tandem_network}, the tandem network architecture consists of an SNN and an ANN that is coupled layer-wise with weights sharing. The ANN is an auxiliary structure that facilitates the error back-propagation for the training of the SNN at the spike-train level, while the SNN is used to derive the exact spiking neural representation. This tandem learning rule allows rapid, efficient, and scalable pattern recognition with SNNs as demonstrated through extensive experimental studies.

The rest of this paper is organized as follows: in Section II, we formulate the proposed tandem learning framework. In Section III, we evaluate the proposed tandem learning framework on both the conventional frame-based vision datasets (i.e., MNIST, CIFAR-10, and ImageNet-12) as well as the event-based vision datasets (i.e., N-MNIST and DVS-CIFAR10) by comparing with other SNN implementations. Finally, we conclude with discussions in Section IV.

\section{Learning Through a Tandem Network}
In this section, we first introduce spiking neuron models that are used in this work. We then present a discrete neural representation scheme using spike count as the information carrier across network layers, and we design ANN activation functions to effectively approximate the spike count of the coupled SNN for error back-propagation at the spike-train level. Finally, we introduce the tandem network and its learning rule, which is called tandem learning rule, for deep SNN training. 
 
\subsection{Neuron Model}
The spiking neuron models describe the rich dynamical behaviors of biological neurons in the brain \cite{gerstner2002spiking}. In general, the computational complexity of spiking neuron models grows with the level of biological plausibility. Therefore, for implementation on efficient neuromorphic hardware, a simple yet effective spiking neuron model that can provide a sufficient level of biological details is preferred.

In this work, we use the arguably simplest spiking neuron models that can effectively describe the sensory information with spike counts: the current-based integrate-and-fire (IF) neuron \cite{ethImageNet} and leaky integrate-and-fire (LIF) neuron models \cite{gerstner2002spiking}. While the IF and LIF neurons do not emulate the rich spectrum of spiking activities of biological neurons, they are however ideal for working with sensory input where information is encoded in spike rates or coincident spike patterns. 

The subthreshold membrane potential ${U_i^l}$ of LIF neuron $i$ at layer $l$ can be described by the following linear differential equation
\begin{equation}
{\tau _{\rm{m}}}\frac{{dU_i^l}}{{dt}} =  - [U_i^l(t) - {U_{rest}}] + RI_i^l(t)
\label{eq:lif_differential}
\end{equation}
where $\tau _{\rm{m}}$ is the membrane time constant. $U_{rest}$ and $R$ are the resting potential and the membrane resistance of the spiking neuron, respectively. $I_i^l(t)$ refers to the time-dependent input current to the neuron $i$. By removing the membrane potential leaky effect involved in the LIF neuron, the subthreshold dynamics of the IF neuron can be described as follows 
\begin{equation}
\frac{{dU_i^l}}{{dt}} = RI_i^l(t)\\
\label{eq:if_differential}
\end{equation}

Without loss of generality, we set the resting potential $U_{rest}$ to zero and the membrane resistance $R$ to unitary in this work. An output spike is generated whenever ${U_i^l}$ crosses the firing threshold $\vartheta$ 
\begin{equation}
s_i^l(t) = \Theta \left(U_i^l(t) - \vartheta\right) \;\; with \;\; \Theta \left(x\right) = \left\{\begin{array}{l}1,\;\;\; if\;x \ge 0\\
0,\;\;\;otherwise\;
\end{array} \right.
\label{eq:spike_generation}
\end{equation}
where $s_i^{l}(t)$ indicates the occurrence of an output spike from the neuron $i$ at time step $t$. 

In practice, given a small simulation time step $dt$, the linear differential equation of the LIF neuron can be well approximated by the following discrete-time formulation
\begin{equation}
U_i^l[t] = \alpha U_i^l[t - 1] + I_i^l[t] - \vartheta s_i^l[t - 1]\;
\label{eq:lif_neuron}
\end{equation}
with 
\begin{equation}
I_i^l[t] = \sum\nolimits_j {w_{ij}^{l - 1}s_j^{l - 1}} [t-1] + b_i^l\\
\label{eq:synaptic_current}
\end{equation}
where $\alpha \equiv \exp \left(-{dt/{\tau _m}} \right)$. The square brackets are used in the above formulations to reflect the discrete-time modeling. $I_i^l[t]$ summarizes the synaptic current contributions from presynaptic neurons of the preceding layer. $w_{ij}^{l - 1}$ denotes the strength of the synaptic connection from the afferent neuron $j$ of layer $l-1$ and $b_i^l$ is the constant injecting current to the neuron $i$. As denoted by the last term of Eq. \ref{eq:lif_neuron}, instead of resetting the membrane potential to zero after each spike generation, the firing threshold $\vartheta$ is subtracted from the membrane potential. This effectively preserves the surplus membrane potential that increased over the firing threshold and reduces the information loss across layers \cite{ethImageNet}. Similarly, the discrete-time formulation of the IF neuron can be expressed as follows
\begin{equation}
U_i^l[t] = U_i^l[t - 1] + I_i^l[t] - \vartheta s_i^l[t - 1]
\label{eq:if_neuron}
\end{equation}

In our experiments, for both IF and LIF neurons, the $U_i^l[0]$ is reset and initialized to zero before processing each new input example. We consider the total number of spikes (i.e., spike count) generated by spiking neurons as the main information carrier. For neuron $i$ at layer $l$, the spike count $c_i^{l}$ can be determined by summing all output spikes over the encoding time window $T$ 
\begin{equation}
c_i^{l} = \sum\nolimits_{t=1}^T {s_i^{l}[t]} 
\label{eq:spike_count}
\end{equation}

In this work, we use the activation value of non-spiking analog neurons to approximate the spike count of spiking neurons. The transformation performed by the analog neuron $i$ can be described as 
\begin{equation}
a_i^l = f(\sum\limits_j {w_{ij}^{l - 1}x_j^{l - 1} + b_i^l})
\label{eq:analog_neuron}
\end{equation}
where $w_{ij}^{l - 1}$ and $b_i^l$ are the weight and bias terms of the analog neuron respectively. $x_j^{l - 1}$ and $a_i^l$ correspond to the analog input and output activation values. $f(\cdot)$ denotes the activation function of analog neurons. Details of the spike count approximation using analog neurons will be explained in Section \ref{discrete_rep}. 

\subsection{Encoding and Decoding Schemes}
The SNNs process inputs that are represented as spike trains, which ideally should be generated by event-based sensors, for instance, silicon retina event camera \cite{6887319} and silicon cochlea audio sensor \cite{6658899}. However, the datasets collected from these event-driven sensors are not abundantly available in comparison to their frame-based counterparts. To take frame-based sensor data as inputs, SNNs will require additional neural encoding mechanisms to transform the real-valued samples into spike trains. 

In general, two neural encoding schemes are commonly considered: rate code and temporal code. Rate code \cite{diehl2015fast,ethImageNet} converts real-valued inputs into spike trains at each sampling time step following a Poisson or Bernoulli distribution. However, it suffers from sampling errors, thereby requiring a long encoding time window to compensate for such errors. Hence, the rate code is not the best to encode information into a short time window that we desire. On the other hand, temporal coding uses the timing of a single spike to encode information. Therefore, it enjoys superior coding efficiency and computational advantages. However, it is complex to decode and sensitive to noise \cite{gerstner2002spiking}. Moreover, it is also challenging to achieve a high temporal resolution, which is essential for the temporal coding, on neuromorphic chips.

Alternatively, we take the real-valued inputs as the time-dependent input currents and directly apply them in Eqs. \ref{eq:lif_neuron} and \ref{eq:if_neuron} at every time step. This neural encoding scheme overcomes the sampling error of the rate code, therefore, it can support the accurate and rapid inference as been demonstrated in earlier works \cite{wu2018direct, bellec2018long}. As shown in Fig. \ref{fig:tandem_network}, beginning from this neural encoding layer, spike trains and spike counts are taken as input to the SNN and ANN layers, respectively.

\begin{figure}[htb]	
	\begin{minipage}[b]{1.0\linewidth}
		\centering
		\centerline
		{\includegraphics[width = 9 cm]{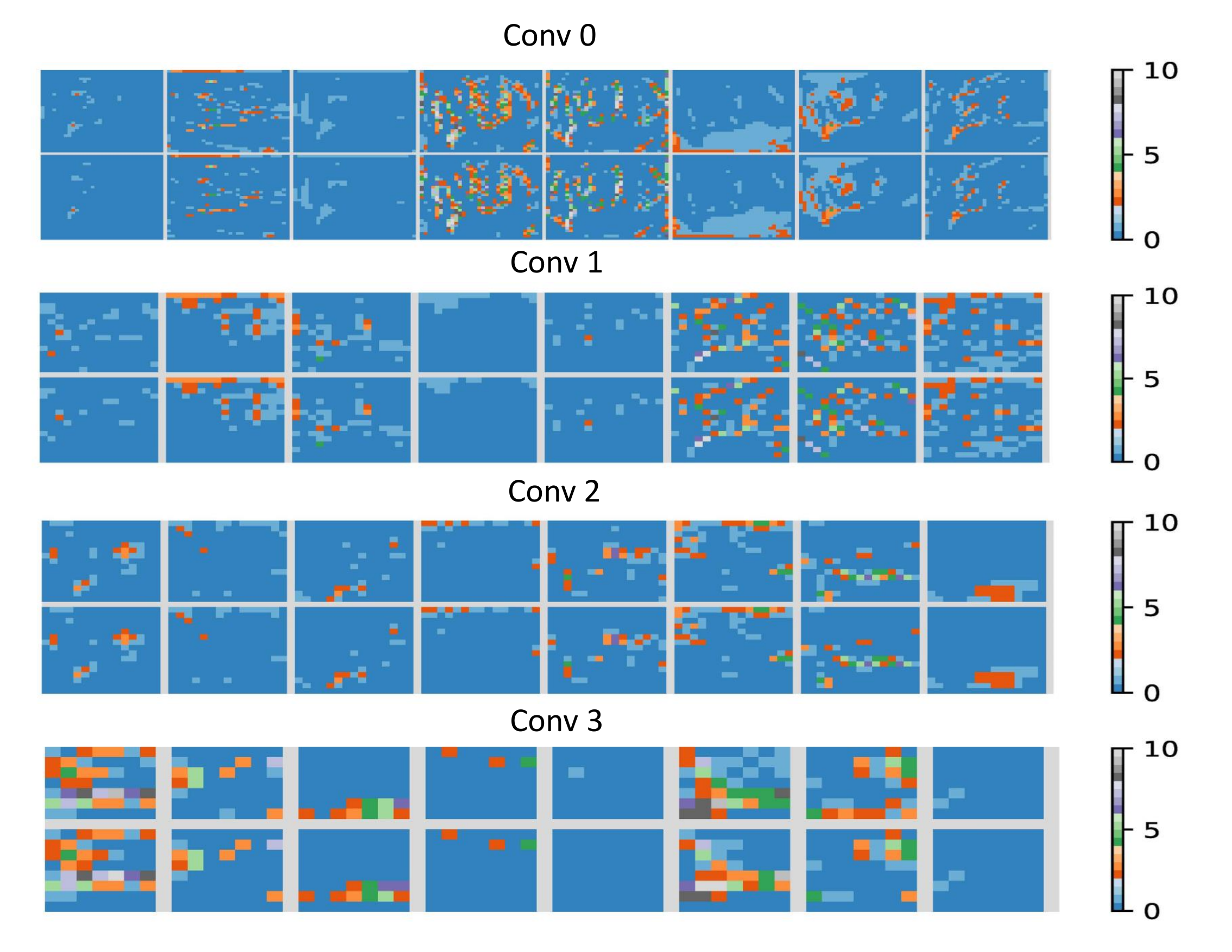}}
		\caption{Illustration of spike counts as the discrete neural representation in the tandem network (IF neurons). The intermediate activations of a randomly selected sample from the CIFAR-10 dataset are provided. The top and bottom row of each convolution layer refer to the exact and approximated spike count activations, derived from the SNN and the coupled ANN layer respectively. Note that only the first 8 feature maps are given and plotted in separated blocks. 
		}
		\label{distRrep}
	\end{minipage}
\end{figure}

To facilitate pattern classification, a SNN back-end is required to decode the output spike trains into pattern classes. For decoding, it is feasible to decode from the SNN output layer using either the discrete spike counts or the continuous free aggregate membrane potentials (no spiking) $U_i^{l,f}$ that accumulated over the encoding time window $T$
\begin{equation}
U_i^{l,f}= R\left(\sum\nolimits_j {w_{ij}^{l - 1} c_j^{l - 1}} + b_i^l T \right)\\
\label{eq:aggregate_mem}
\end{equation}

In our preliminary study, as shown in Fig. \ref{learningcurve}, we observe that the free aggregate membrane potential provides a much smoother learning curve, as it allows continuous error gradients to be derived at the output layer. Furthermore, the free aggregate membrane potential can be directly used as the output for regression tasks. Therefore, we use the free aggregate membrane potential for neural decoding in this work unless otherwise stated.

\subsection{Spike Count as a Discrete Neural Representation}
\label{discrete_rep}
Deep ANNs learn to describe the input data with compact latent representations. A typical latent representation is in the form of a continuous or discrete-valued vector. While most studies have focused on continuous latent representations, discrete representations have their unique advantages in solving real-world problems \cite{van2017neural, mnih2014neural, salakhutdinov2009deep, mnih2016variational, courville2011spike}. For example, they are potentially a more natural fit for representing natural language which is inherently discrete, and also native for logical reasoning and predictive learning. Moreover, the idea of discrete neural representation has also been exploited in the network quantization \cite{courbariaux2016binarized,jacob2018quantization}, where network weights, activation values, and gradients are quantized for efficient neural network training and inference. 

In this work, we consider the spike count as a discrete latent representation in deep SNNs and design ANN activation functions to  approximate the spike count of the coupled SNN, such that spike-train level surrogate gradients can be effectively derived from the ANN layer. With such a discrete latent representation, the effective non-linear transformation at the SNN layer can be expressed as
\begin{equation}
c_i^l = g({s^{l - 1}};w_i^{l - 1},b_i^l)
\end{equation}
where $g(\cdot)$ denotes the effective neural transformation performed by spiking neurons. Given the state-dependent nature of spike generation, it is not feasible to directly determine an analytical expression from $s^{l - 1}$ to $c_i^l$. To circumvent this problem, we simplify the spike generation process by assuming the resulting synaptic currents from $s^{l - 1}$ are evenly distributed over time. This yields a constant synaptic current $I_i^{l,c}$ at every time step
\begin{equation}
I_i^{l,c} \equiv \left( {\sum\nolimits_j {w_{ij}^{l - 1}c_j^{l - 1} + b_i^lT} } \right)/T
\end{equation}

Taking the constant synaptic current $I_i^{l,c}$ into Eq. \ref{eq:if_differential}, we thus obtain the following expression for the interspike interval of IF neurons 
\begin{equation}
ISI_i^l = \rho \left( {\frac{\vartheta }{RI_i^{l,c}}} \right) 
\end{equation}
where $\rho(\cdot)$ denotes the non-linear transformation of the Rectified Linear Unit (ReLU). As mentioned in the earlier section, the membrane resistance R is assumed to be unitary in this work and hence dropped. The output spike count can be further approximated as follows
\begin{equation}
c_i^l = \frac{{{T}}}{{ISI_i^l}} = \frac{1}{\vartheta } \;\rho \left({\sum\nolimits_j {w_{ij}^{l - 1}c_j^{l - 1} + b_i^l {T}}} \right)
\label{eq:sc_if}
\end{equation}

\begin{figure}[!htb]	
	\centering
	\centerline
	{\includegraphics[width = 9 cm]{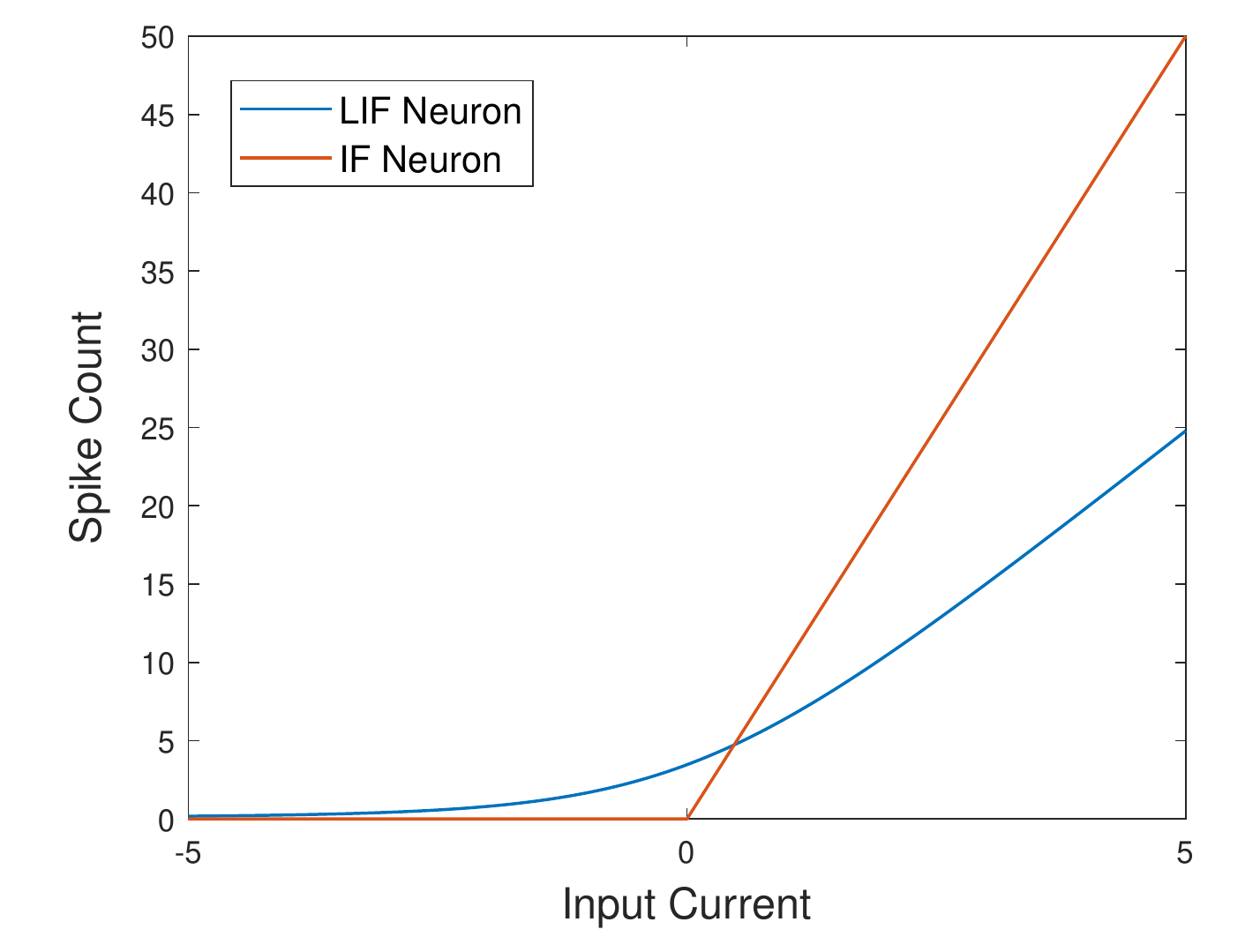}}
	\caption{Illustration of the neuronal activation functions that are designed to approximate the spike count of the IF and LIF neurons. In this example, a firing threshold of 1 and 0.1 is used for IF and LIF neurons respectively, and a encoding time window of 10 is considered. The membrane time constant $\tau _m$ is set to 20 time steps for the LIF neuron.}
	\label{fig:surrogate_activation_function}
\end{figure}

\begin{figure*}[htb]	
	\centering
	\centerline
	{\includegraphics[width = 16 cm]{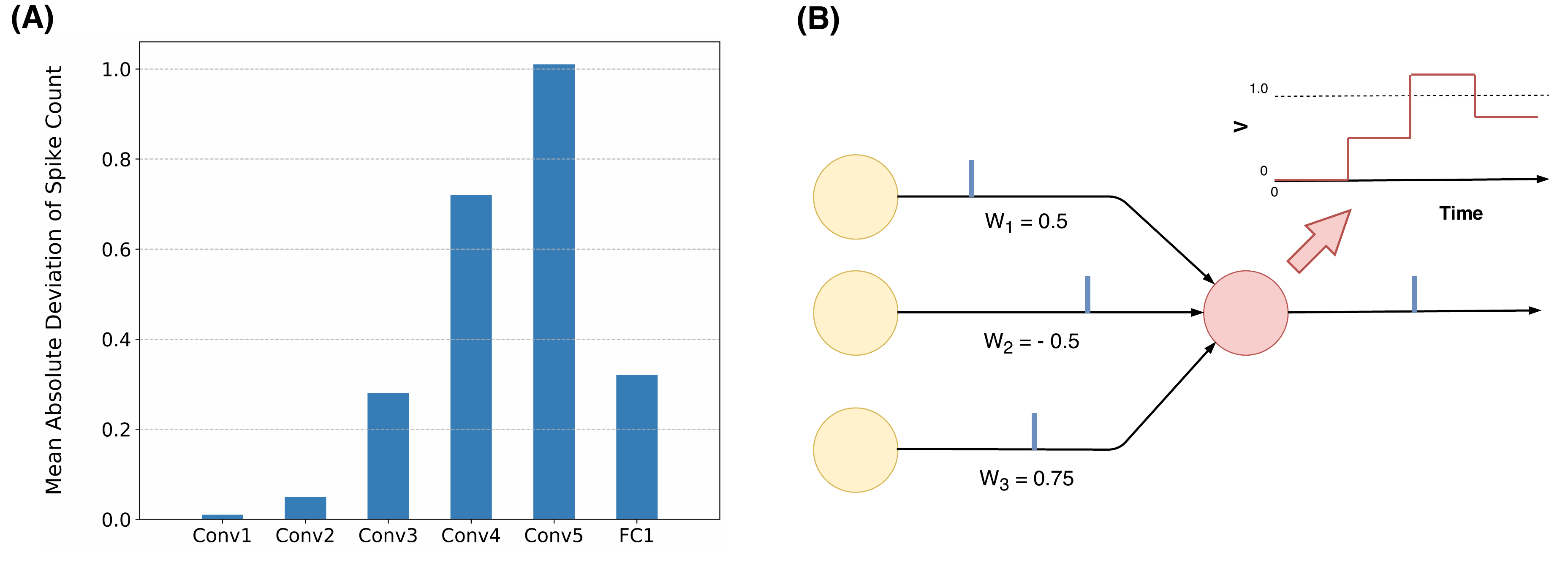}}
	\caption{(A) Summary of the neural representation error that happened with the constrain-then-train approach, i.e., the mean spike count difference between the actual SNN layer outputs and those approximated from a constrained ANN of the same weights. The experiment is performed with a network structure of CifarNet (IF neurons) and an encoding time window of 10. (B) A hand-crafted example for illustration of the spike count approximation error between the SNN and the approximated ANN, which usually happens when the encoding time window is short and neuronal activities are sparse. In this example, although the aggregate membrane potential of the postsynaptic IF neuron stays below the firing threshold in the end (a useful intermediate quantity that is applied to approximate the output spike count), an output spike is generated due to the early arrival of spikes from excitatory synapses.}
	\label{fig:spike_count_mismatch}
\end{figure*}

By setting $\vartheta$ to 1, Eq. \ref{eq:sc_if} takes the same form as the activation function of analog neurons as described in Eq. \ref{eq:analog_neuron}. Furthermore, by using $\rho(\cdot)$ as the activation function for analog neurons, spike count $c_j^{l - 1}$ as the input, and aggregated constant injecting current $b_i^l T$ as the bias term for the corresponding analog neurons, this configuration allows the spike count and hence the spike-train level error gradients to be approximated from the coupled weight sharing ANN layer. As shown in Fig. \ref{distRrep}, it is apparent that the proposed ANN activation function can effectively approximate the exact spike count of the coupled SNN layers in an image classification task. The approximation errors can be considered as stochastic noise that was shown to improve the generalizability of the trained neural networks \cite{noh2017regularizing}.

Following the same approximation mechanism for the IF neuron, one also consider injecting a constant current into the LIF neuron, the interspike interval can be determined from Eq. \ref{eq:lif_differential} by calculating the charging duration for neurons to rise from the resting potential to the firing threshold. Thus, we obtain 
\begin{equation}
ISI_{\rm{i}}^l = {\tau _m}\log \left[ {1 + \frac{\vartheta }{{\rho (I_i^{l,c} - \vartheta )}}} \right]
\end{equation}

Hence, the approximated spike count can be evaluated as
\begin{equation}
c_i^l = \frac{T}{{{\tau _m}}}\log {\left[ {1 + \frac{\vartheta }{{\rho (I_i^{l,c} - \vartheta )}}} \right]^{ - 1}}
\label{eq:sc_lif}
\end{equation}

However, the above equation is undefined when $I_i^{l,c} \le \vartheta $, and it is also numerically unstable when $I_i^{l,c} - \vartheta $ is marginally greater than zero. To address this, we replace the ReLU activation function $\rho (\cdot)$ with a smoothed surrogate ${\rho _s}(\cdot) $ that is defined as follows,
\begin{equation}
{\rho _s}(x) = \log (1 + {e^x})
\end{equation}

Same as the IF neurons, by taking the spike count $c_j^{l - 1}$ and the aggregated constant injecting current  $b_i^l T$ as inputs for analog neurons,  with Eq. \ref{eq:sc_lif} as the activation function, the spike count of the LIF neurons can be well approximated by the coupled analog neurons. The two activation functions for IF and LIF neurons are shown in Fig. \ref{fig:surrogate_activation_function}. The discrete neural representation with spike count and its approximation with weight sharing analog neurons allow the surrogate gradients to be approximated at the spike-train level and applied during error back-propagation. Hence, it has a learning efficiency superior to other surrogate gradient methods \cite{neftci2019surrogate, shrestha2018slayer, wu2018direct}, which take place at each time step.

\subsection{Credit Assignment in the Tandem Network}
As the customized ANN activation functions can effectively approximate the discrete neural representation of spiking neurons, it prompts us to think whether it is feasible to directly train a constrained ANN and then transfer its weights to an equivalent SNN, i.e., the constrain-then-train approach \cite{hunsberger2016training, wu2019deep}. In this way, the training of a deep SNN can be achieved by that of a deep ANN, and a large number of tools and methods developed for ANNs can be leveraged.

\begin{figure}[!htb]	
	\centering
	\centerline
	{\includegraphics[width = 9 cm]{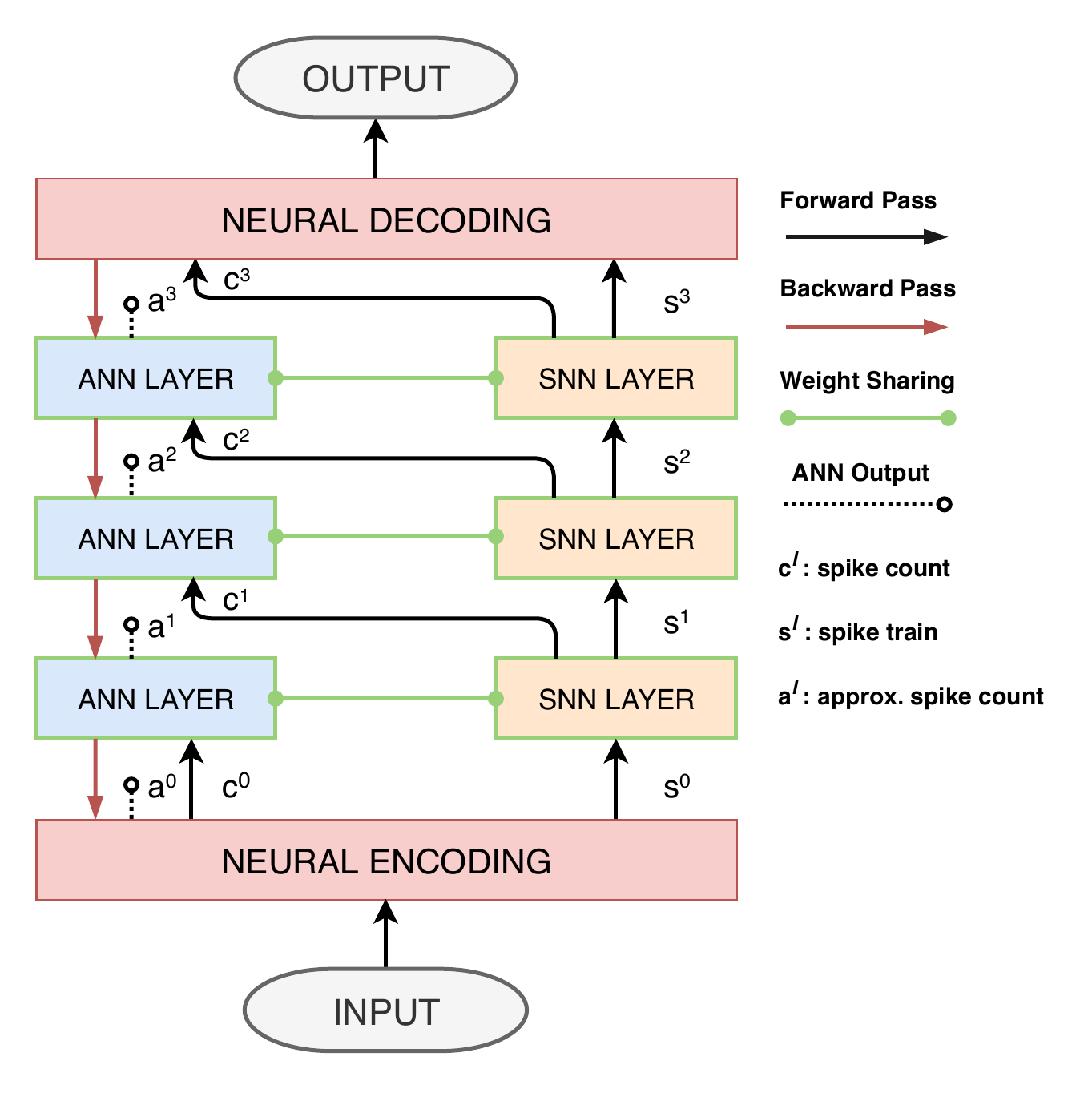}}
	\caption{Illustration of the proposed tandem learning framework that consists of an SNN and an ANN with shared weights. The spike counts are considered as the main information carrier in this framework. ANN activation function is designed to approximate the spike counts of the coupled SNN, so as to approximate the gradients of the coupled SNN layers at the spike-train level. During training, in the forward pass, the synchronized spike counts and spike trains derived from an SNN layer are taken as the inputs to the subsequent SNN and ANN layers respectively; the error gradients are passed backward through the ANN layers during error back-propagation, to update the weights so as to minimize the objective function.}
	\label{fig:tandem_network}
\end{figure}

We take  Eq. \ref{eq:sc_if} as the activation function for a constrained ANN, and subsequently transfer the trained weights to the SNN with IF neurons. The resulting network reports a competitive classification accuracy on the MNIST dataset\cite{wu2019deep}. However, when applying this approach to the more complex CIFAR-10 dataset with a time window of 10, a large classification accuracy drop (around 21\%) occurred to the SNN from that of the pre-trained ANN. By carefully comparing the ANN approximated `spike count' with the actual SNN spike count, we observe an increasing spike count discrepancy between the same layers of ANN and SNN as shown in Fig. \ref{fig:spike_count_mismatch}A. The discrepancy grows as the spike counts travel through the layers. This is due to the fact that the activation function of the ANN has ignored the temporal dynamics of the IF neuron. While such spike count discrepancies could be negligible for a shallow network used for classifying the MNIST dataset\cite{wu2019deep} or with a very long encoding time window \cite{hunsberger2016training}, it has huge impacts in the face of sparse synaptic activities and short time window. 

To demonstrate how such an approximation error may occur during information forward-propagation, namely neural representation error, we hand-craft an example as shown in Fig. \ref{fig:spike_count_mismatch}B. Although the free aggregate membrane potential, at the end of the encoding time window, of an IF neuron stays below the firing threshold, an output spike could have been generated due to the early arrival of spikes from the excitatory synapses. We understand that the spike count, approximated by the surrogate activation function in the analog neuron, only provides a mean estimation, which ignores the temporal jitter of spike trains. It is worth mentioning that the spike count discrepancy can be well controlled at the single layer as shown in Fig. \ref{distRrep} and Fig. \ref{fig:spike_count_mismatch}A (the discrepancy in the first layer is insignificant). However, such a neural representation error will accumulate across layers and significantly affect the classification accuracy of the SNN with weights transferred from a trained ANN. Therefore, to effectively train a deep SNN with a short encoding time window and sparse synaptic activities, it is necessary to derive an exact neural representation with SNN in the training loop. 

To solve this problem, we propose a tandem learning framework. As shown in Fig. \ref{fig:tandem_network}, an ANN with activation function defined in Eqs. \ref{eq:sc_if} and \ref{eq:sc_lif} are employed to enable error back-propagation through the ANN layers; while the SNN, sharing weights with the coupled ANN is employed to determine the exact neural representation (i.e., spike counts and spike trains). The synchronized spike counts and spike trains, determined from the SNN layer, are transmitted to the subsequent ANN and SNN layers, respectively. It is worth mentioning that, in the forward pass, the ANN layer takes the output of the previous SNN layer as the input. This aims at synchronizing the inputs of the SNN with ANN via the interlaced layers, rather than trying to optimize the classification performance of the ANN.  

\begin{algorithm}
	\small
	\DontPrintSemicolon
	\KwInput{Input sample $X_{in}$, target label $Y$, parameters $w$ of a $L$-layer network, encoding time window size $T$}
	\KwOutput{Updated network parameters $w$}
	\vspace{4mm}
	\KwFw{}       
	${c}^0, {s}^0$ = Encoding($X_{in}$) \\
	\For{layer $l = 1$ to $L-1$}
	{
		\tcp*[l]{\scriptsize State Update of the ANN Layer} 
		$a^l$ = ANN.layer[$l$].forward($c^{l-1}$, $w^{l - 1}$) $^{*}$ \\
		\For{t = 1 to $T$}
		{\tcp*[l]{\scriptsize State Update of the SNN Layer} 
			$s^{l}[t]$ = SNN.layer[$l$].forward($s^{l-1}[t]$, $w^{l - 1}$)  
		}
		\tcp*[l]{\scriptsize Update the Spike Count}
		$c^{l}$ =  $\sum\nolimits_{t=1}^{T} {s^{l}[t]}$ 
	} 
	\tcc{\scriptsize Output with Different Decoding Schemes}
	\If{Decode with `Aggregate Membrane Potential'}
	{
		$output$ = ANN.layer[$l$].forward($c^{L-1}$, $w^{L-1}$)
	}
	\ElseIf{Decode with `Spike Count'}
	{
		\For{t = 1 to $T$}
		{
			$s^{L}[t]$ = SNN.layer[$l$].forward($s^{L-1}[t]$, $w^{L-1}$)\\
		}
		$output = \sum\nolimits_{t=1}^{T} {s^{L}[t]}$
	}
	\KwLoss{$E$ = LossFunction($Y, output$)} 
	\vspace{4mm}
	
	\KwBw{}
	$\frac{{\partial E}}{{\partial a^{L}}}$ = LossGradient($Y, output$) \\
	\For{layer $l = L-1$ to $1$}
	{	
		\tcp*[l]{\footnotesize Gradient Update through the ANN Layer} 
		$\frac{{\partial E}}{{\partial {a^{l - 1}}}}$ , $\frac{{\partial E}}{{\partial {w^{l - 1}}}}$ = ANN.layer[$l$].backward($\frac{{\partial E}}{{\partial {a^{l}}}}$ , $c^{l-1}$, $w^{l - 1}$) 
	}
	\vspace{2mm}
	Update parameters of the ANN layer based on the calculated gradients. \\
	Share the updated parameters with the coupled SNN layer. \\
	\vspace{2mm}
	\KwNote{}
	$^{*}$ For inference, state updates are performed on the SNN layers entirely.
	\caption{Pseudo Codes of the Tandem Learning Rule}
	\label{algo}
\end{algorithm}

\begin{figure*}[htb]	
	\centering
	\centerline
	{\includegraphics[width = 16 cm]{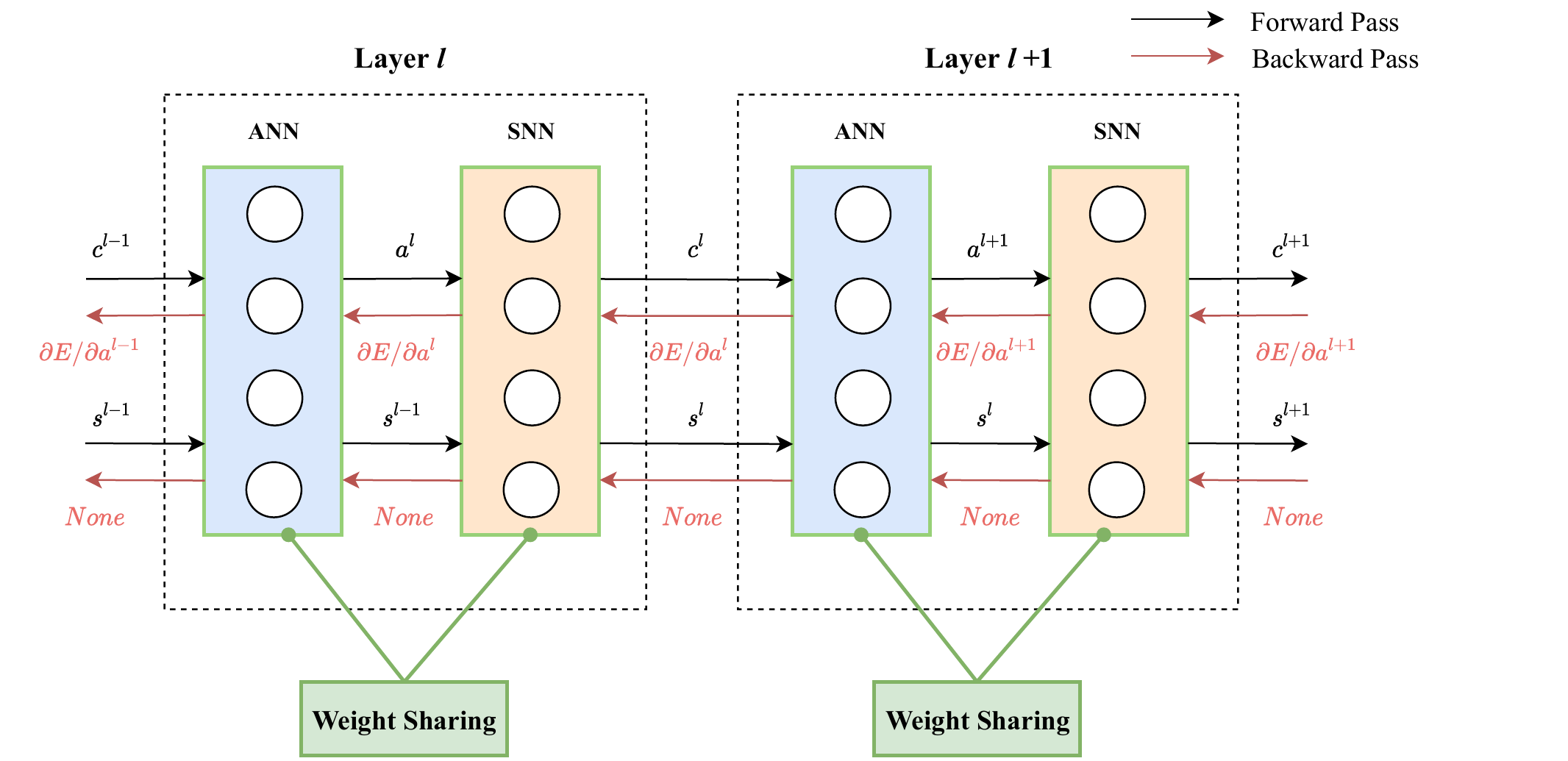}}
	\caption{Illustration of the tandem network implementation in Pytorch and Tensorpack, where the coupled ANN and SNN layers (i.e., convolution and fully-connected layers) are encapsulated into a customized module that can be imported conveniently to construct tandem networks. Within each module, the ANN and SNN layers are connected in tandem to allow the appropriate information to be propagated through the network. During forward pass, the SNN layer takes the input spike train from the preceding layer to determine the exact output spike counts and spike trains of the layer. During the backward pass, the SNN layer works effectively as a bridge to transmit the received gradient information to the coupled ANN layer, such that the spike-train level gradients can be approximated by the coupled ANN layer.}
	\label{fig:tandem_implementation}
\end{figure*}

By incorporating the dynamics of spiking neurons during the training of the tandem network, the exact output spike counts, instead of ANN predicted spike counts, are propagated forward to the subsequent ANN layer. The proposed tandem learning framework can effectively prevent the neural representation error from accumulating forward across layers. While a coupled ANN is harnessed for error back-propagation, the forward inference is executed entirely on the SNN after training. The pseudo code of the proposed tandem learning rule is given in Algorithm \ref{algo}. 

\section{Experimental Evaluation and Discussion}
In this section, we first evaluate the learning capability of the proposed tandem learning framework on frame-based object recognition and image reconstruction tasks. We further discuss why effective learning can be performed within the tandem network. Then, we evaluate the applicability of the tandem learning rule to the asynchronous inputs generated from the event-driven camera sensors. Finally, we discuss the properties of high learning efficiency and scalability, rapid inference as well as synaptic operation reductions that can be achieved with the proposed tandem learning rule. 

\subsection{Experimental Setups}
\paragraph{Datasets}
To conduct a comprehensive evaluation on the proposed tandem learning rule, we use three conventional frame-based image datasets: MNIST, CIFAR-10 \cite{krizhevsky2009learning}, and ImageNet-12 \cite{deng2009imagenet}. The MNIST dataset consists of gray-scaled handwritten digits of 28$\times$28 pixels, including 60,000 training and 10,000 testing samples. The CIFAR-10 consists of 60,000 color images of size 32$\times$32$\times$3 from 10 classes, with a standard split of 50,000 and 10,000 for training and testing, respectively. The ImageNet-12 dataset consists of over 1.2 million images from 1,000 object categories, which remains a challenging benchmark for deep SNN models. 

In addition, we also investigate the applicability of tandem learning to event-driven vision datasets: N-MNIST \cite{10.3389/fnins.2015.00437} and DVS-CIFAR10 \cite{10.3389/fnins.2017.00309}. The N-MNIST dataset is collected by moving the event-driven camera in front of an LCD monitor that displays samples from the frame-based MNIST dataset. The event-driven camera  mounted on an actuated pan-tilt camera platform follows three microsaccades for each sample, and each of these microsaccades takes 100 ms. In contrast, the DVS-CIFAR10 dataset is collected by fixing the event-driven camera, while moving the image on the monitor over four paths that takes a total of 200 ms. The total number of samples reduced to one-sixth for the collected DVS-CIFAR10 dataset due to the huge size of storage required. Following the similar data pre-processing procedures adopted in \cite{wu2018direct}, we reduce the temporal resolution by accumulating the spikes occurred within every 10 ms intervals for the N-MNIST and DVS-CIFAR10 datasets.

\paragraph{Network and Training Configurations}
As shown in Table. \ref{tab:networks}, we use a convolutional neural network (CNN) with 7 learnable layers for object recognition on both the frame-based CIFAR-10 and event-based DVS-CIFAR10 datasets, namely CifarNet. To handle the higher input dimensionality of the DVS-CIFAR10, we increase the stride of each convolution layer to 2 and the kernel size to 7 for the first layer. Due to the high computational cost and large memory requirements for discrete-time modeling of SNNs, we use the AlexNet \cite{krizhevsky2012imagenet} for object recognition on the large-scale ImageNet-12 dataset. For object recognition on the N-MNIST dataset, we design a 7-layer CNN that called the DigitNet. For image reconstruction task on the MNIST dataset, we evaluate on a spiking autoencoder that has an architecture of 784-256-128-64-128-256-784, wherein the numbers refer to the number of neurons at each layer.

To reduce the dependency on weight initialization and to accelerate the training process, we add batch normalization layer after each convolution and fully-connected layer. Given that the batch normalization layer only performs an affine transformation, we follow the approach introduced in \cite{ethImageNet} and integrate their parameters into the preceding layer's weights before applying them in the coupled SNN layer. We replace the average pooling operations, that are commonly used in ANN-to-SNN conversion works, with a stride of 2 convolution operations, which not only perform dimensionality reduction in a learnable fashion but also reduce the computation cost and latency \cite{zhang2018shufflenet}. 

For SNNs with IF neurons, we set the firing threshold to 1. For SNNs with LIF neurons, we set the firing threshold to 0.1 and the membrane time constant $\tau _m$ to 20 time steps. The corresponding ANN activation functions are provided in Fig. \ref{fig:surrogate_activation_function}.

\paragraph{Implementation Details}
\begin{table}[ht]
	\centering
	\small
	\resizebox{9cm}{!}{
		\begin{tabular}{|p{1.0cm}|p{2.00cm}|p{1.0 cm}|p{2.00cm}|p{1.0 cm}|p{2.00cm}|}
			\hline
			\multicolumn{2}{|c|}{(A) CifarNet } & \multicolumn{2}{c|}{(B) AlexNet} & \multicolumn{2}{c|}{(C) DigitNet} \\
			\hline
			Conv1 & (3, 3, 1, 128) & Conv1 & (12, 12, 4, 96) & Conv1 & (3, 3, 1, 32) \\
			Conv2 & (3, 3, 2, 256) & Conv2 & (5, 5, 2, 256) & Conv2 & (3, 3, 2, 64)\\
			Conv3 & (3, 3, 2, 512) & Conv3 & (3, 3, 2, 384) & Conv3 & (3, 3, 2, 64)\\
			Conv4 & (3, 3, 1, 1024) & Conv4 & (3, 3, 1, 384) & Conv4 & (3, 3, 2, 128)\\
			Conv5 & (3, 3, 1, 512) & Conv5 & (3, 3, 2, 256) & Conv5 & (3, 3, 1, 256)\\
			FC1 & (1024) & FC1 & (4096) & FC1 & (1024)\\
			Dropout & Prob=0.2 & FC2 & (4096) & Dropout & Prob=0.2\\
			FC2 & (10) & FC3 & (1000) & FC2 & (10)\\
			\hline
		\end{tabular}
	}
	\caption{Network architectures used for the (A) CIFAR-10 and DVS-CIFAR10, (B) ImageNet-2012, and (C) N-MNIST experiments. For Conv layers, the values in the bracket refer to the height, width, stride, and number of filters, respectively. For the FC layer, the value in the bracket refers to the number of neurons.}
	\label{tab:networks}
\end{table} 

We perform all experiments with Pytorch\cite{paszke2019pytorch}, except for the experiment on the ImageNet-12 dataset where we use the Tensorpack toolbox \cite{wu2016tensorpack}. Tensorpack is a high-level neural network training interface based on the TensorFlow, which optimizes the whole training pipeline for the ImageNet-12 object recognition task and provide accelerated and memory-efficient training on multi-GPU machines. We follow the same data pre-processing procedures (crop, flip and mean normalization, etc.), optimizer, learning rate decay schedule that are adopted in the Tensorpack for ImageNet-12 object recognition task. For the object recognition task on the CIFAR-10, we follow the same data pre-processing and training configurations as  in \cite{wu2018direct}. For the ease of tandem network implementations, as shown in Fig. \ref{fig:tandem_implementation}, we encapsulate the coupled ANN and SNN layers (convolution and fully-connected layers) into customized modules that can be imported conveniently to construct tandem networks in Pytorch and Tensorpack. 

\paragraph{Evaluation Metrics}
\label{count_synops}
For object recognition tasks on both the frame-based and event-based vision datasets, we report the classification accuracy on the test sets. For the image reconstruction task on the MNIST dataset, we report the mean square error of the reconstructed handwritten digits. We perform 3 independent runs for all tasks and report the best result across all runs, except for the object recognition task on the ImageNet-12 dataset where the performance of only a single run is reported.

To study the computational efficiency of SNN models over their ANN counterparts, we follow the convention of neuromorphic computing community by counting the total synaptic operations \cite{merolla2014million,ethImageNet}. For SNN, as defined below, the total synaptic operations (SynOps) correlate with the neurons' firing rate, fan-out $f_{out}$ (number of outgoing connections to the subsequent layer), and time window size $T$.
\begin{equation}
SynOps = \sum\limits_{t = 1}^{T} {\sum\limits_{l = 1}^{L-1} {\sum\limits_{j = 1}^{N^l} {f_{out,j}^l} } } s_j^l[t]
\end{equation}
where $L$ is the total number of layers and $N^l$ denotes the total number of neurons in layer $l$. 

In contrast, the total synaptic operations that are required to classify one image in the ANN is given as follows
\begin{equation}
SynOps = \sum\limits_{l = 1}^L {{f_{in}^l}}{N^l}
\end{equation}
where $f_{in}^l$ denotes the number of incoming connections to each neuron in layer $l$. Given a particular network structure, the total $SynOps$ is fixed for the ANN. In our experiment, we calculate the average synaptic operations on a randomly chosen mini-batch (256 samples) from the test set, and report the SynOps ratio between SNN and ANN.

\begin{figure*}[htb]	
	\centering
	\centerline
	{\includegraphics[width = 18 cm]{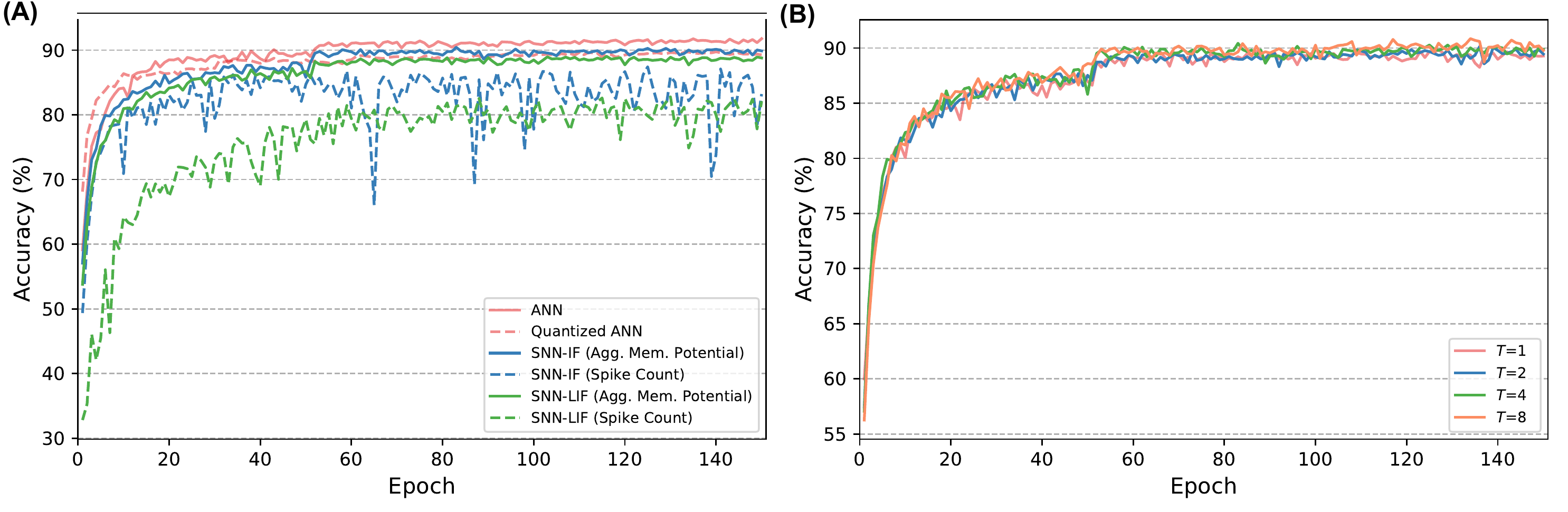}}
	\caption{(A) Classification accuracy on the CIFAR-10 test set with different training schemes. (B) Classification accuracy on the CIFAR-10 test set as a function of different encoding window sizes $T$. The IF neurons are used in this experiment.}
	\label{learningcurve}
\end{figure*}

\begin{table*}[!htb]
	\centering	
	\caption{Comparison of classification accuracy and inference speed of different SNN implementations on the CIFAR-10 and ImageNet-12 test sets.  } 				
	\resizebox{18 cm}{!}{
		\begin{tabular}{llcccc}
			\toprule
			\toprule 
			&\textbf{Model} & \textbf{Network Architecture} &\textbf{Method} & \textbf{Error Rate (\%)} & \textbf{Inference Time Steps}\\ 
			\toprule 
			\parbox[t]{2mm}{\multirow{10}{*}{\rotatebox[origin=c]{90}{\textbf{CIFAR-10}}}}&Panda and Roy (2016)\cite{panda2016unsupervised} & Convolutional Autoencoder & Layer-wise Spike-based Learning & 75.42 & - \\
			&Hunsberger and Eliasmith (2016)\cite{hunsberger2016training}  & AlexNet & Constrain-then-Train  & 83.54 & 200\\ 	
			&Rueckauer et al. (2017)\cite{ethImageNet} & 8-layer CNN & ANN-to-SNN conversion  & 90.85 & -\\	
			&Sengupta et al. (2019)\cite{sengupta2019going} & VGG-16 & ANN-to-SNN conversion & 91.46 & 2,500\\  
			&Lee et al. (2019)\cite{lee2019enabling} & ResNet-11 & ANN-to-SNN conversion & 91.59 & 3,000\\
			&Lee et al. (2019)\cite{lee2019enabling} & ResNet-11 & Spike-based Learning & 90.95 & 100 \\ 					
			&Wu et al. (2019)\cite{wu2018direct}  & 8-layer CNN & Surrogate Gradient Learning & 90.53 &-\\							
			&Wu et al. (2019)\cite{wu2018direct}  & AlexNet & Surrogate Gradient Learning & 85.24 & -\\	
			&\textbf{This work (ANN with full-precision activation)} & CifarNet & Error Backpropagation &\textbf{91.77} & \textbf{-} \\
			&\textbf{This work (ANN with quantized activation)} & CifarNet & Error Backpropagation &\textbf{89.99} & \textbf{-} \\								
			&\textbf{This work (SNN-IF with Spike Count)} & CifarNet & Tandem Learning &\textbf{87.41} & \textbf{8}\\
			&\textbf{This work (SNN-IF with Agg. Mem. Potential)} & CifarNet & Tandem Learning &\textbf{90.98} & \textbf{8} \\
			&\textbf{This work (SNN-LIF with Spike Count)} & CifarNet & Tandem Learning &\textbf{82.78} & \textbf{8}\\
			&\textbf{This work (SNN-LIF with Agg. Mem. Potential)} & CifarNet & Tandem Learning &\textbf{89.04} & \textbf{8} \\
			\toprule						
			\toprule
			\parbox[t]{2mm}{\multirow{6}{*}{\rotatebox[origin=c]{90}{\textbf{ImageNet}}}}
			&Rueckauer et al. (2017)\cite{ethImageNet}  & VGG-16 & ANN-to-SNN conversion  & 49.61 (81.63) & 400 \\						
			&Sengupta et al. (2019)\cite{sengupta2019going}  & VGG-16 & ANN-to-SNN conversion  & 69.96 (89.01) & 2,500\\
			&Hunsberger and Eliasmith (2016)\cite{hunsberger2016training}  & AlexNet & Constrain-then-Train  & 51.80 (76.20) & 200\\ 	
			&\textbf{This work (ANN with full-precision activation)} & AlexNet & Error Backpropagation &\textbf{57.55 (80.44)} & \textbf{-} \\
			&\textbf{This work (ANN with quantized activation)} & AlexNet & Error Backpropagation &\textbf{50.27 (73.92)} & \textbf{-} \\										
			&\textbf{This work (SNN-IF with Agg. Mem. Potential)} & AlexNet & Tandem Learning &\textbf{46.63 (70.80)} & \textbf{5} \\
			&\textbf{This work (SNN-IF with Agg. Mem. Potential)} & AlexNet & Tandem Learning &\textbf{50.22 (73.60)} & \textbf{10} \\
			\toprule
			\bottomrule
		\end{tabular}
		\label{results}
	}
\end{table*}
 
\subsection{Frame-based Object Recognition Results} 
For CIFAR-10, as provided in Table. \ref{results}, the SNN using IF neurons, denoted as SNN-IF hereafter, achieves a test accuracy of 87.41\% and 90.98\% for spike count and aggregate membrane potential decoding, respectively. While the result is slightly worse for the SNN implementation with LIF neurons (SNN-LIF) that achieves a classification accuracy of 89.04\%, which may be due to the approximation error of the smoothed surrogate activation function. Nevertheless, with a similar activation function designed to approximate the firing rate of LIF neurons \cite{hunsberger2016training}, the constrain-then-train approach only achieves a classification accuracy of 83.54\% on the CIFAR-10 dataset. This result confirms the necessity of keeping the SNN in the training loop as proposed in the tandem learning network. Moreover, the results achieved by our spiking CifarNet is also as competitive as the state-of-the-art ANN-to-SNN conversion \cite{ethImageNet,sengupta2019going,lee2019enabling} and spike-based learning \cite{lee2019enabling,wu2018direct} methods.

As shown in Fig. \ref{learningcurve}A, we note that the learning dynamics with spike count decoding is unstable, which is attributed to the discrete error gradients derived at the output layer. Therefore, we use the aggregate membrane potential decoding for the rest of the experiments. Although the learning converges slower than the plain CNN with the ReLU activation function and quantized CNN (with activation value quantized to 3 bits following the quantization-aware training scheme \cite{jacob2018quantization}), the classification accuracy of the SNN-IF eventually surpasses that of the quantized CNN. To study the effect of the encoding time window size $T$ on the classification accuracy, we repeat the CIFAR-10 experiments using IF neurons with $T$ ranging from 1 to 8. As shown in Fig. \ref{learningcurve}B, the classification accuracy improves consistently with a larger time window size. It suggests the effectiveness of tandem learning in making use of the encoding time window, which determines the upper bound of the spike count, to represent information. Notably, 89.83\% accuracy can be achieved with a time window size of only 1, suggesting accurate and rapid inference can be achieved simultaneously.

To train a model on the large-scale ImageNet-12 dataset with a spike-based learning rule, it requires a huge amount of computer memory to store the intermediate states of the spiking neurons as well as huge computational costs for a discrete-time simulation. Hence, only a few SNN implementations, without taking into consideration the dynamics of spiking neurons during training, have made some successful attempts on this challenging task, including ANN-to-SNN conversion \cite{ethImageNet,sengupta2019going} and constrain-then-train \cite{hunsberger2016training} approaches. 

\begin{figure*}[htb]	
	\centering
	\centerline
	{\includegraphics[width = 16 cm]{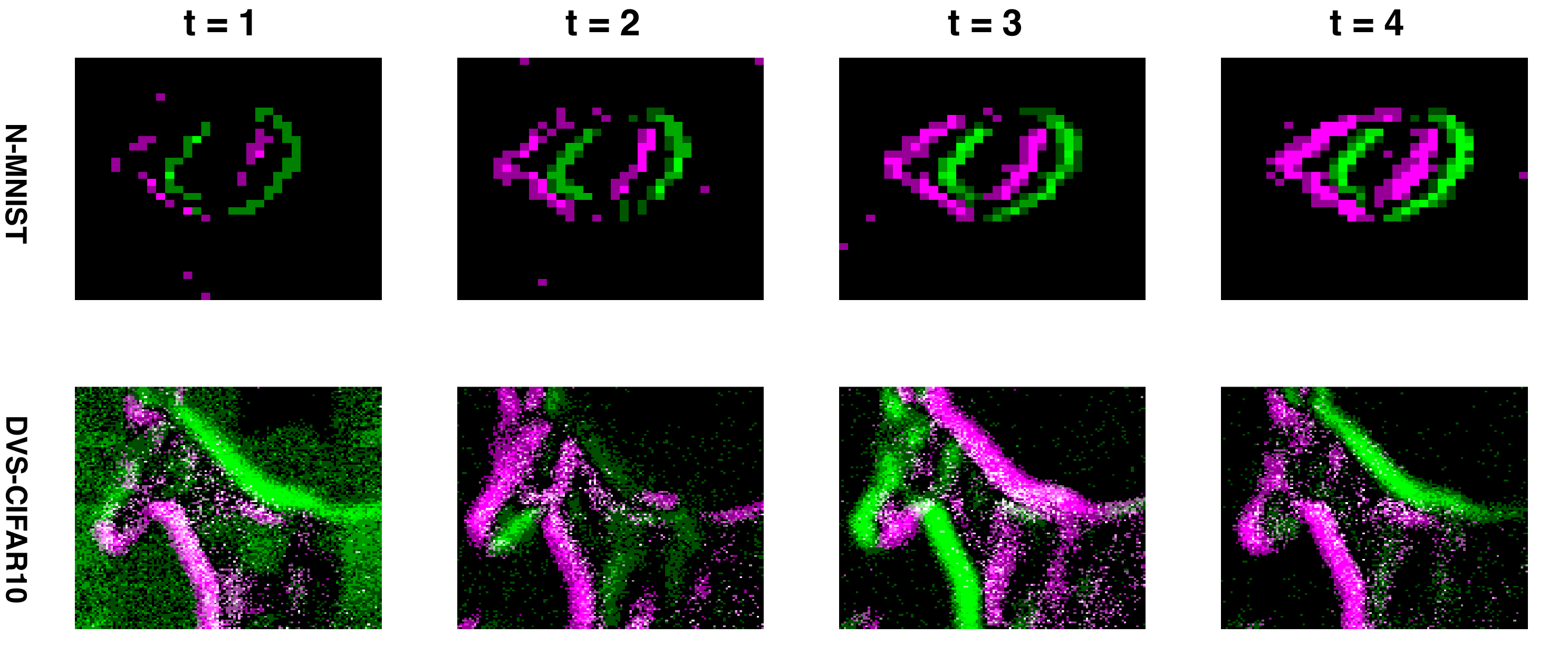}}
	\caption{Illustration of samples collected from the event-based camera. The samples are constructed by aggregating spiking events that occurred within each 10 ms interval. The `On' and `Off' events at each pixel are color-coded with `purple' and `green', respectively.}
	\label{fig:neuromorphic_data}
\end{figure*}

\begin{table*}[htb]
	\centering				
	\caption{Comparison of the object recognition results on the N-MNIST and DVS-CIFAR10 datasets.} 				
	\resizebox{11 cm}{!}{
		\begin{tabular}{lccc}
			\toprule
			\toprule 
			\textbf{Model} & \textbf{Method} & \multicolumn{2}{c}{\textbf{Accuracy}}\\ 
			\cline{3-4} \\
			\textbf{} & \textbf{} & N-MNIST & DVS-CIFAR10\\ 
			\toprule 	
			Neil, Pfeiffer and Liu \cite{neil2016phased} & Phased LSTM & 97.30\% & - \\
			Lee, Delbruck and Pfeiffer \cite{lee2016training} & Fully-connected SNN & 98.78\% & - \\
			Jin et al. \cite{jin2018hybrid} & Fully-connected SNN & 98.88\% & - \\			
			Li et al. \cite{li2017cifar10} & BOE-Random Forest & - & 31.01\% \\ 
			Orchard et al. \cite{orchard2015hfirst} & H-First & 71.20 \% & 7.7\% \\
			Lagorce et al. \cite{lagorce2016hots} & HOTS & 80.80\% & 27.1\% \\ 
			Sironi et al. \cite{sironi2018hats} & Gabor-SNN & 83.70\% & 24.50\% \\								
			Sironi et al. \cite{sironi2018hats} & HATS & 99.10\% & 52.4\% \\
			Wu et al. \cite{wu2018direct}& Spiking CNN & 99.35\% & 58.10\% \\ 																
			Shrestha and Orchard \cite{shrestha2018slayer}& Spiking CNN & 99.20\% & -\\ 																																																																																								
			\textbf{This work (IF)} & Spiking CNN & \textbf{99.31\%} & \textbf{58.65\%} \\ 					
			\textbf{This work (LIF)} & Spiking CNN & \textbf{99.22\%} & \textbf{57.18\%} \\ 	
			\textbf{This work (IF) + Fine-tuning} & Spiking CNN & \textbf{-} & \textbf{65.59\%} \\ 	
			\textbf{This work (LIF) + Fine-tuning} & Spiking CNN & \textbf{-} & \textbf{63.73\%} \\ 								
			\toprule
			\bottomrule
		\end{tabular}
	}
	\label{table:event_vision}
\end{table*}

As shown in Table. \ref{results}, with an encoding time window of 10 time steps, the AlexNet trained with the tandem learning rule achieves top-1 and top-5 accuracies of 50.22\% and 73.60\%, respectively. This result is comparable to that of the constrain-then-train approach with the same AlexNet architecture \cite{hunsberger2016training} with a total number of time steps of 200. Notably, the proposed learning rule only takes 10 inference time steps which are at least an order of magnitude faster than the other reported methods. While the ANN-to-SNN conversion approaches \cite{ethImageNet,sengupta2019going} achieve better classification accuracies on the ImageNet-12, their successes are largely credited to the more advanced network models used. 

Furthermore, we note tandem learning suffers an accuracy drop of around 7\% from the baseline ANN implementation with full-precision activation (revised from the original AlexNet model \cite{krizhevsky2012imagenet} by replacing pooling layers with a stride of 2 convolution operations to match the AlexNet used in this work and adding batch normalization layers). To investigate the effect on the accuracy of the discrete neural representation (how much of the drop in the accuracy is due to activation quantization, and how much of it is due to the dynamics of the IF neuron), we modify the full-precision ANN by quantizing the activation values to only 10 levels. In a single trial, the resulting quantized ANN achieves the top-1 and top-5 error rate of 50.27\% and 73.92\%, respectively. This result is very close to that of our SNN implementation, which suggests that the quantization of the activation function alone accounts for most of the accuracy drop.

\subsection{Event-based Object Recognition Results} 
The bio-inspired event cameras capture per-pixel intensity change asynchronously which exhibits compelling properties of high dynamic range, high temporal resolution, and no motion blur. Event-driven vision \cite{gallego2019event} therefore attracts growing attention in the computer vision community as a complement to the conventional frame-based vision. The early research on event-driven vision focuses on constructing frame-based feature representation from the captured event streams, such that it can be effectively processed by machine learning models \cite{lagorce2016hots,sironi2018hats} or artificial neural networks \cite{gehrig2019end}. Despite the promising results achieved by these works, the post-processing of frame-based features increases latency and incurs high computational cost even during low event rates. In contrast, the asynchronous SNNs naturally process event-based sensory inputs and hence hold great potential to build fully event-driven systems.

In this work, we investigate the applicability of tandem learning in training SNNs to handle inputs of event camera. To this end, we perform object recognition tasks on the N-MNIST and DVS-CIFAR10 datasets (examples from these datasets are provided in Fig. \ref{fig:neuromorphic_data}). As shown in Table. \ref{table:event_vision}, for the N-MNIST dataset, our spiking CNNs achieve an accuracy of 99.31\% and 99.22\% for SNN-IF and SNN-LIF, respectively. These results outperform many existing SNN implementations \cite{lee2016training,jin2018hybrid,li2017cifar10,shrestha2018slayer} and machine learning models \cite{neil2016phased,lagorce2016hots,sironi2018hats}, while on par with the best reported result achieved in a recently introduced spike-based learning method \cite{wu2018direct}. 

Similarly, our SNNs models also report state-of-the-art performance on the DVS-CIFAR10 dataset. This demonstrates the effectiveness of the proposed tandem learning rule in handling event-driven camera data. To address the data scarcity of the DVS-CIFAR10 dataset, we further explored transfer learning by fine-tuning SNN models (pre-trained on frame-based CIFAR10 dataset), on the DVS-CIFAR10 dataset. Notably, the SNN models trained in this way achieve approximately 7\% accuracy improvements. It is worth noting that, despite neglecting the temporal structure of spike trains and only consider spike counts as the information carrier, the tandem learning rule performs exceedingly well on these datasets, which can be explained by the fact that negligible temporal information is added during the collection of these datasets \cite{iyer2018neuromorphic}.

\subsection{Superior Regression Capability} 
\begin{figure}[htb]	
	\centering
	\centerline
	{\includegraphics[width = 8.5 cm]{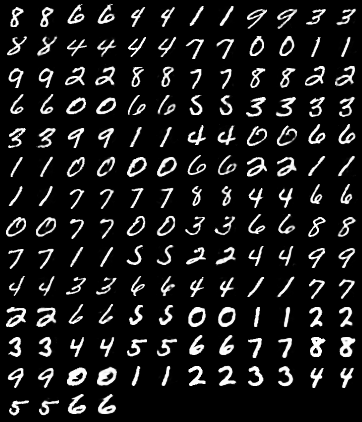}}
	\caption{Illustration of the reconstructed images from a spiking autoencoder (IF neurons with $T$=32) on the MNIST dataset. For each digit, the left column is the original image and the right column is the reconstructed image.}
	\label{fig:ae_demo}
\end{figure}

\begin{figure}[htb]	
	\centering
	\centerline
	{\includegraphics[width = 8 cm]{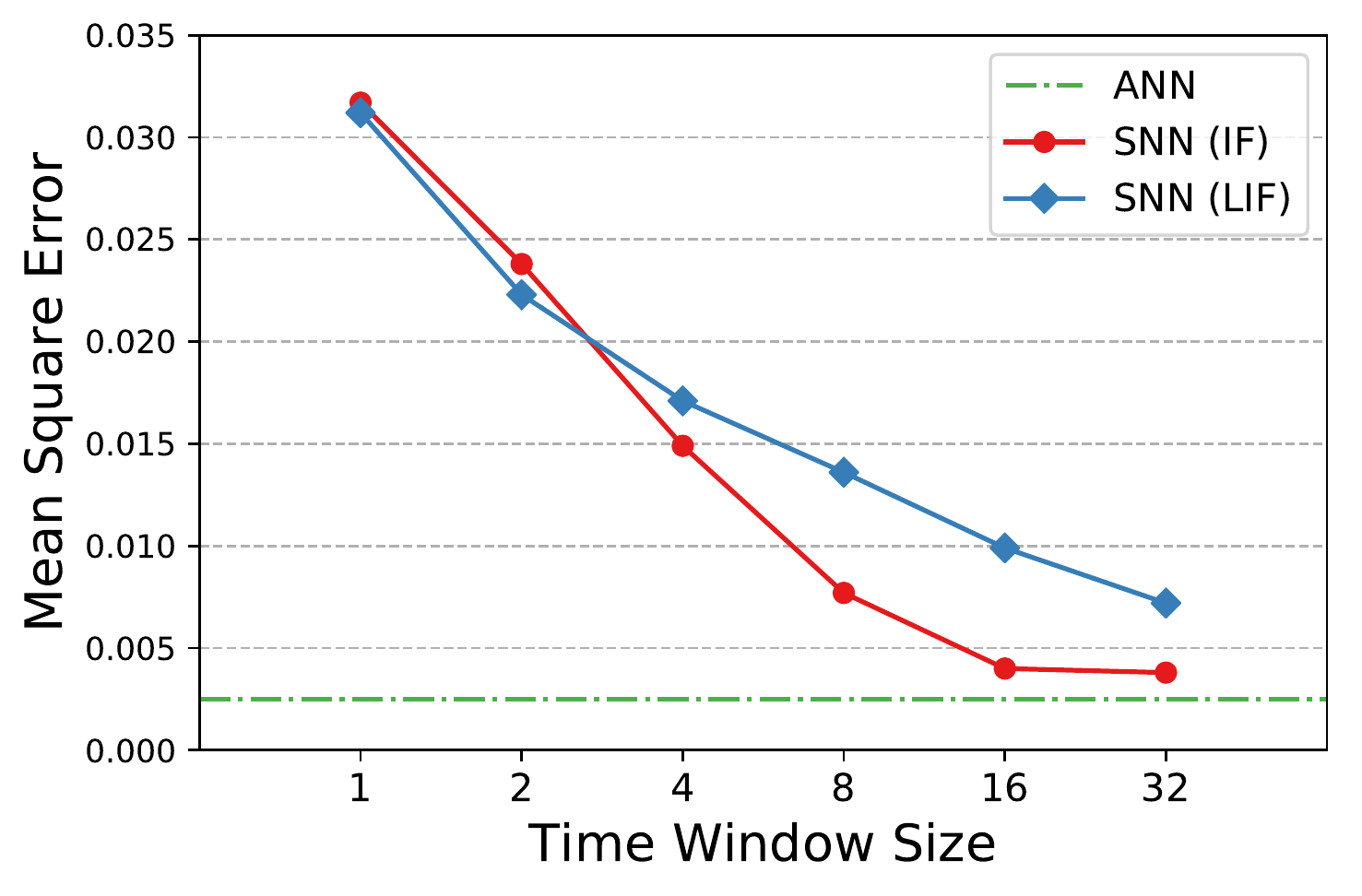}}
	\caption{Illustration of the image reconstruction performance as a function of the encoding time window size.}
	\label{fig:ae_result_vs_time}
\end{figure}

To explore the capability of tandem learning for regression tasks, we perform an image reconstruction task on the MNIST dataset with a fully-connected spiking autoencoder. As shown in Fig. \ref{fig:ae_demo}, the autoencoder trained with the proposed tandem learning rule can effectively reconstruct images with high quality. With a time window size of 32, the SNN-IF and SNN-LIF achieve mean square errors (MSE) of 0.0038 and 0.0072, a slight drop from 0.0025 of an equivalent ANN. However, it is worth mentioning that by leveraging the sparsity of spike trains, the SNNs can provide a much higher data compression rate over a high-precision floating number representation of the ANN. As shown in Fig. \ref{fig:ae_result_vs_time}, with a larger time window size, the network performance approaches that of the baseline ANN, which aligns with the observation in object recognition tasks.

\begin{figure*}[htb]	
	\centering
	\centerline
	{\includegraphics[width = 18 cm]{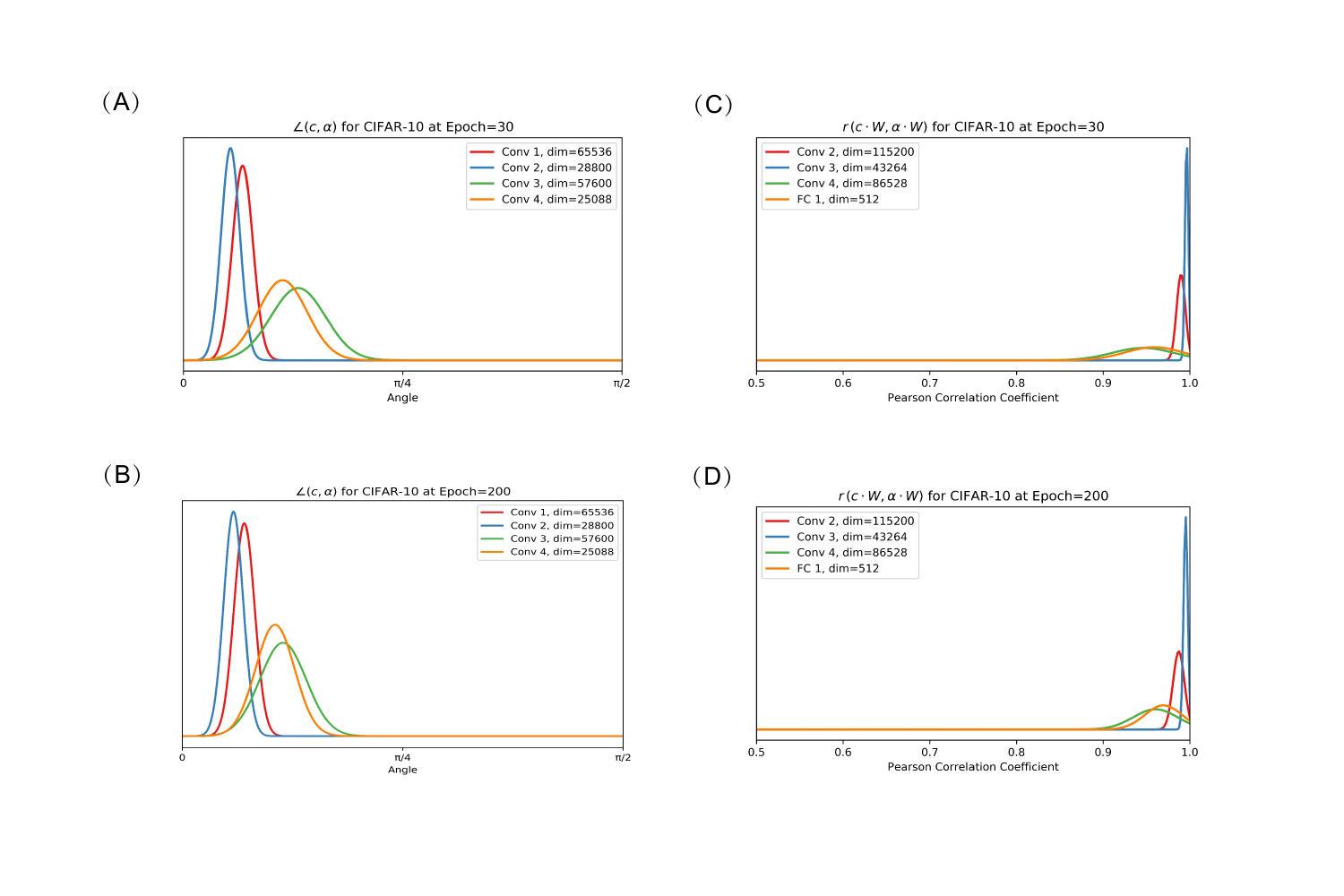}}
	\caption{Analysis of mismatch errors between output spike counts of ANN and SNN layers. Distribution of cosine angles between vectorized $c$ and $a$ for all convolution layers at Epoch 30 (A) and 200 (B). While these angles seem large in low dimensions, they are exceedingly small in a high dimensional space. Distribution of Pearson Correlation Coefficients between weight-activation dot products $c \cdot W$ and $a \cdot W$ at Epoch 30 (C) and 200 (D). The Pearson Correlation Coefficients maintain consistently above 0.9 throughout learning, suggesting that the linear relationship of weight-activation dot products are approximately preserved. }
	\label{angle}
\end{figure*}

\subsection{Activation Direction Preservation and Weight-Activation Dot Product Proportionality within the Interlaced Layers}

After showing how effective the proposed tandem learning rule performs on object recognition and image reconstruction tasks, we hope to explain why the learning can be performed effectively via the interlaced network layers. To answer this question, we borrow ideas from a recent theoretical work of binary neural network \cite{anderson2017high}, wherein learning is also performed across the interlaced network layers (binarized activations are forward propagated to subsequent layers). In the proposed tandem network, the ANN layer activation value $a^{l-1}$ at layer $l-1$ is replaced with the spike count $c^{l-1}$ derived from the coupled SNN layer. We further analyze the degree of mismatch between these two quantities and its effect on the activation forward propagation and error back-propagation. 

In our numerical experiments on CIFAR-10 with a randomly drawn mini-batch of 256 test samples, we calculate the cosine angle between vectorized $c^{l-1}$ and $a^{l-1}$ for all the convolution layers. As shown in Fig. \ref{angle}, their cosine angles are below 24 degrees on average and such a relationship maintains consistently throughout learning. While these angles seem large in low dimensions, they are exceedingly small in a high dimensional space. According to the hyperdimensional computing theory \cite{kanerva2009hyperdimensional} and the theoretical study of binary neural network \cite{anderson2017high}, the cosine angle between any two high dimensional random vectors is approximately orthogonal. It is also worth noting that the distortion of replacing $a^{l-1}$ with $c^{l-1}$ is less severe than binarizing a random high dimensional vector, which changes cosine angle by 37 degrees in theory. Given that the activation function and error gradients that back-propagated from the subsequent ANN layer remains equal, the distortions to the error back-propagation are bounded locally by the discrepancy between $a^{l-1}$ and
$c^{l-1}$.

Furthermore, we calculate the Pearson Correlation Coefficient (PCC)  between the weight-activation dot products $c^l \cdot W$ and $a^l \cdot W$, which is an important intermediate quantity (input to the batch normalization layer) in our current network configurations. The PCC, ranging from -1 to 1, measures the linear correlation between two variables. A value of 1 implies a perfect positive linear relationship. As shown in Fig. \ref{angle}, the PCC maintains consistently above 0.9 throughout learning for most of the samples, suggesting the linear relationship of weight-activation dot products are approximately preserved. 

\subsection{Efficient Learning through Spike-Train Level Surrogate Gradient}
In this section, we compare the learning efficiency of the proposed tandem learning rule to the popular family of surrogate gradient learning methods \cite{neftci2019surrogate}. The surrogate gradient learning methods describe the time-dependent dynamic of spiking neurons with a recurrent neural network, whereby the BPTT-based training method is used to optimize the network parameters. The non-differentiable spike generation function is replaced by a continuous surrogate function during the error back-propagation phase, such that a surrogate gradient can be determined for each time step. In contrast, the tandem learning determines the error gradient at the spike-train level, therefore, it can significantly improve the learning efficiency.

\begin{figure*}[htb]	
	\centering
	\centerline
	{\includegraphics[width = 16 cm]{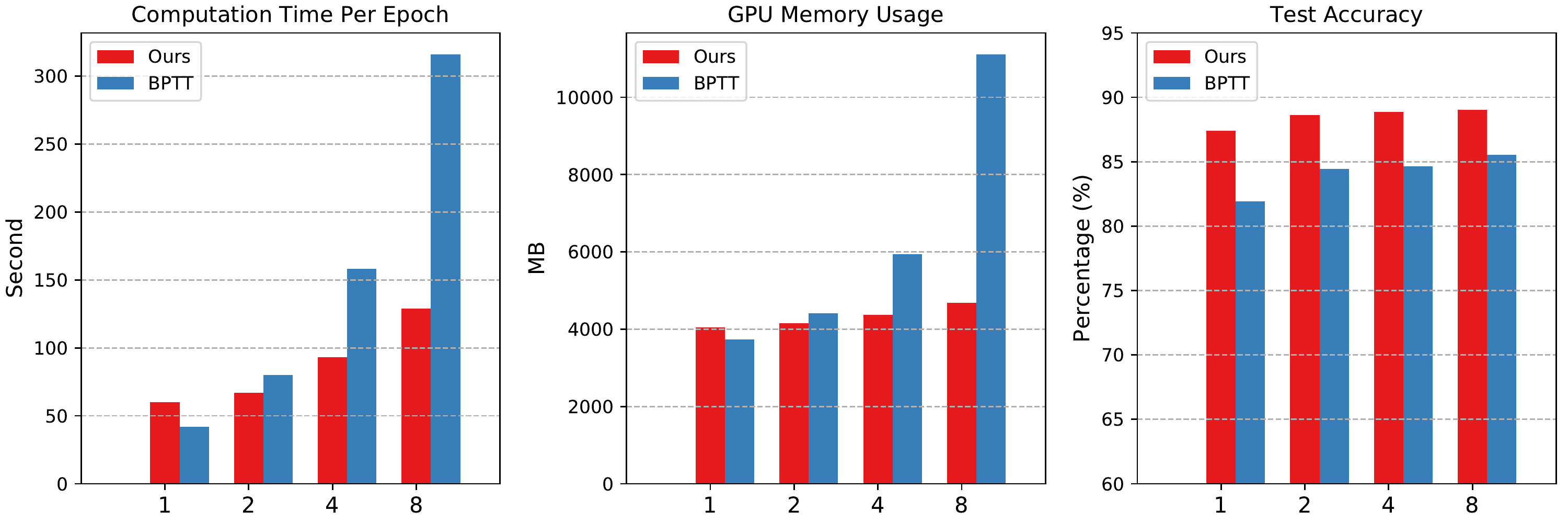}}
	\caption{Comparison of the computation time, GPU memory usage and test accuracy of the proposed tandem learning approach over the BPTT-based surrogate gradient method. The results are provided as a function of the encoding time window size $T$.}
	\label{compute_efficiency}
\end{figure*}

Here, we compare the learning efficiency of tandem learning with the surrogate learning method presented in \cite{wu2018direct}. As shown in Fig. \ref{compute_efficiency}, for the experiment on the CIFAR-10 dataset with LIF neurons, the computation time and GPU memory usage grow linearly with the time window size $T$ for the BPTT-based method, since it requires to store and calculate using the intermediate neuronal states. It worth noting that, taking this BPTT-based method, the SNN is unable to fit onto a single Nvidia Geforce GTX 1080Ti GPU card with 11 GB memory space when $T=8$. Therefore, it prevents a large-scale deployment of this approach for more challenging tasks as also mentioned in other works \cite{neftci2019surrogate}. In contrast, the storage of the intermediate neuronal state of each time step is not required for the tandem learning, therefore, it shows a speed-up of 2.45 time over the BPTT-based approach with 2.37 times fewer GPU memory usage at $T=8$. The improvements in computational efficiency are expected to be further boosted for larger time window sizes. Furthermore, the SNNs trained with the tandem learning approach achieved higher test accuracies over the BPTT-based approach consistently for all different $T$. Therefore, the proposed tandem learning approach demonstrates much better learning effectiveness, efficiency, and scalability.

\subsection{Rapid Inference with Reduced Synaptic Operations}
As shown in Table. \ref{results}, the SNN trained with the proposed learning rule can perform inference at least one order of magnitude quicker than other learning rules without compromising on the classification accuracy. Moreover, as demonstrated in Figs. \ref{learningcurve} and \ref{fig:ae_result_vs_time}, the proposed tandem learning rule can deal with and utilize different encoding time window size $T$. In the most challenging case when only 1 spike is allowed to transmit (i.e., $T=1$), we are able to achieve satisfying results for both object recognition and image reconstruction tasks. This may be partially credited to the encoding scheme that we have employed, whereby full input information can be encoded in the first time step. Besides, the Batch Normalization layer, which is added after each convolution, and fully-connected layer ensure effective information transmission to the top layers. The results can be improved further by increasing the time window size, therefore, a trade-off between inference speed and accuracy can be achieved according to different application requirements. 

\begin{figure*}[htb]    
	\centering
	\centerline
	{\includegraphics[width = 16 cm]{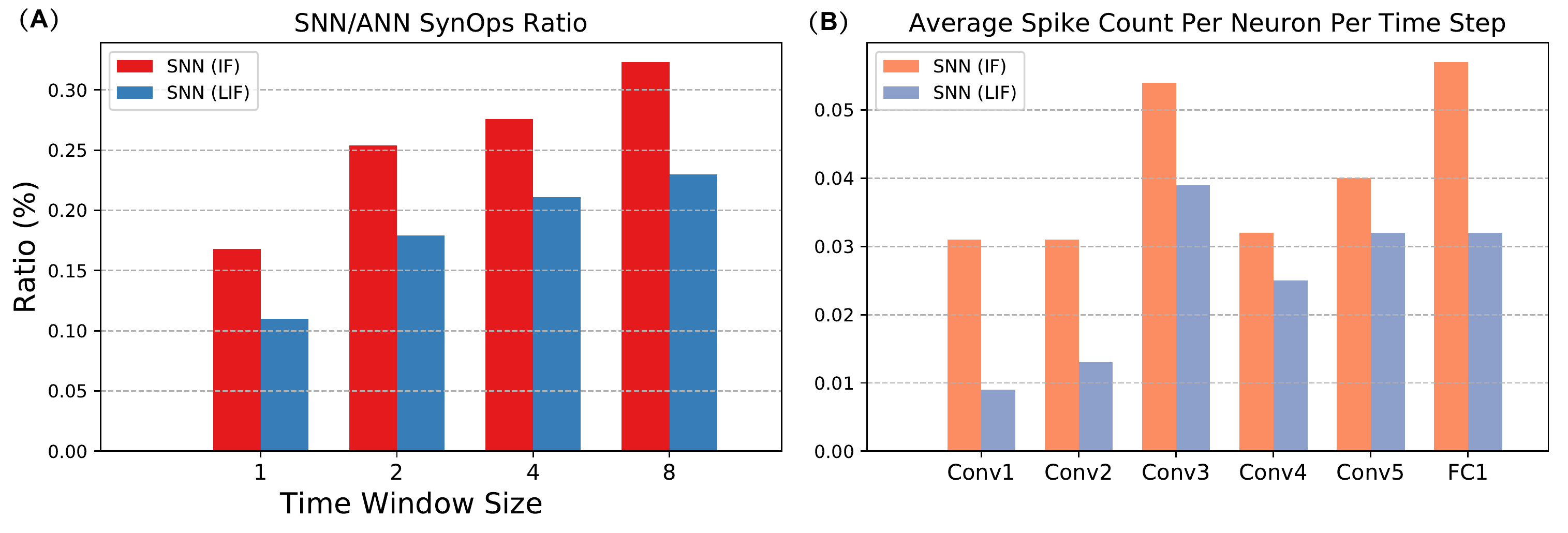}}
	\caption{(A) The total SNN SynOps to ANN SynOps as a function of encoding time window size on the CIFAR-10 dataset. (B) Average spike count per neuron per time step of the trained SNN model ($T=8$). Sparse neuronal activities can be observed across all network layers, leading to low power consumption when implemented on the neuromorphic hardware.}
	\label{avgSpikeCount}
\end{figure*}

To study the energy efficiency of the train SNNs, we follow the convention by calculating the ratio of total SNN SynOps to ANN SynOps on the CIFAR-10 dataset. To achieve a state-of-the-art classification accuracy on the CIFAR-10 dataset ($T=8$), the SNN-IF and SNN-LIF require a SynOps ratio of only 0.32 and 0.22 to the ANN counterpart, respectively. This can be explained by the short inference time required and the sparse synaptic activities as summarized in Fig. \ref{avgSpikeCount}B. Notably, the total SynOps required for ANN is a fixed number that is independent of the time window size, while the total SynOps required for SNN grows almost linearly with time window size as shown in Fig. \ref{avgSpikeCount}A. It is worth noting that the SNN is more energy-efficient than its ANN counterpart when such a ratio is below 1. The saving is even more significant compared to ANNs if we consider the fact that for SNNs, only an accumulate (AC) operation is performed for each synaptic operation; while for ANNs, a more costly multiply-and-accumulate (MAC) operation is performed. This results in an order of magnitude chip area as well as energy-saving per synaptic operation\cite{ethImageNet,sengupta2019going}. In contrast, the state-of-the-art SNN implementations with the ANN-to-SNN conversion and spike-based learning methods require a SynOps ratio of 25.60 and 3.61 respectively on a similar VGGNet-9 network \cite{lee2019enabling}. It suggests our SNN implementation is at least an order of magnitude more efficient.

\section{Discussion and Conclusion}
In this work, we introduce a novel tandem neural network and its learning rule to effectively train SNNs for classification and regression tasks. Within the tandem neural network, an SNN is employed to determine exact spike counts and spike trains for the activation forward propagation; while an ANN, sharing the weight with the coupled SNN, is used to approximate the spike counts and hence gradients of the coupled SNN at the spike-train level. Given that error back-propagation is performed on the simplified ANN, the proposed learning rule is both memory and computationally more efficient than the popular surrogate gradient learning methods that perform gradient approximation at each time step \cite{neftci2019surrogate, zenke2018superspike, shrestha2018slayer, wu2018direct}. It is noted that a similar strategy has been explored for the hardware in-the-loop training on BrainScaleS systems \cite{7966125} to counteract the noises induced by the analog substrate. While the tandem learning approach introduced here addresses the training effectiveness and efficiency of surrogate gradient learning methods. 

To understand why the learning can be effectively performed within the tandem learning framework, we study the learning dynamics of the tandem network and compare it with an intact ANN. The empirical study on the CIFAR-10 reveals that the cosine distances between the vectorized ANN output $a^l$ and the coupled SNN output spike count $c^l$ are exceedingly small in a high dimensional space and such a relationship maintains throughout the training. Furthermore, strongly positive Pearson Correlation Coefficients are exhibited between weight-activation dot product $c^l \cdot W$ and $a^l \cdot W$, an important intermediate quantity in the activation forward propagation, suggesting a linear relationship of weight-activation dot products is well preserved. 

The SNNs trained with the proposed tandem learning rule have demonstrated competitive classification accuracies on both the frame-based and event-based object recognition tasks. By making efficient use of the time window size, that determines the upper bound for the spike count, to represent information, and adding batch normalization layers to ensure effective information flow; rapid inferences, with at least an order of magnitude time-saving compared to other SNN implementations are demonstrated in our experiments. Furthermore, by leveraging on the sparse neuronal activities and short encoding time window, the total synaptic operations are also reduced by at least an order of magnitude over the baseline ANNs and other state-of-the-art SNN implementations. By integrating the algorithmic power of the proposed tandem learning rule with the unprecedented energy efficiency of emerging neuromorphic computing architectures, we expect to enable low-power on-chip computing on pervasive mobile and embedded devices. 

For future work, we will explore strategies to close the accuracy gap between the baseline ANN and SNN using the LIF neurons by designing a more effective approximating function for the LIF neuron as well as to evaluate more advanced network architectures. In addition, we would like to acknowledge that the tandem learning rule, which neglects the temporal structure of spike trains, is not applicable for temporal sequence learning because the error function is required to be determined for each time step or spike rather than at the spike count level. To solve the tasks where the temporal structure is important, such as gesture recognition, we are interested to study whether a hybrid network structure that includes a feedforward network for feature extraction, and a recurrent network for sequence modeling could be useful. Specifically, the feedforward SNN trained with tandem learning can work as a powerful rate-based feature extractor on the short time scale, while a subsequent spiking recurrent neural network \cite{bellec2018long} can be used to explicitly handle the temporal structure of underlying patterns on the longer time scale.

\bibliographystyle{IEEEbib}
\bibliography{citation-main}
\end{document}